\documentclass[letterpaper, 10 pt, conference]{ieeeconf}  

\IEEEoverridecommandlockouts                              

\usepackage{graphicx} 
\usepackage{tabularx}
\usepackage{caption}
\usepackage{subcaption}
\usepackage{lipsum}
\usepackage{soul,color}
\usepackage{amsmath}

\title{\LARGE \bf
Agonist-Antagonist Pouch Motors: Bidirectional Soft Actuators Enhanced by Thermally Responsive Peltier Elements
}

\author{Trevor Exley,  Rashmi Wijesundara, Nathan Tan, Akshay Sunkara, Xinyu He, Shuopu Wang,\\ Bonnie Chan, Aditya Jain, Luis Espinosa, and Amir Jafari
\thanks{*This work was funded by National Science Foundation NSF under Grant Number 2045177, and by the National Institute of Health (NIH) through grant T32GM136501.}
\thanks{$^{1}$All authors are with Advanced Robotic Manipulators (ARM) Lab, the Department of Biomedical Engineering, University of North Texas, Texas, United States.
        {Corresponding author: \tt\small amir.jafari@unt.edu}}%
        }

\begin{document}

\maketitle
\thispagestyle{empty}
\pagestyle{empty}

\begin{abstract}
In this study, we introduce a novel Mylar-based pouch motor design that leverages the reversible actuation capabilities of Peltier junctions to enable agonist-antagonist muscle mimicry in soft robotics. Addressing the limitations of traditional silicone-based materials, such as leakage and phase-change fluid degradation, our pouch motors filled with Novec 7000 provide a durable and leak-proof solution for geometric modeling. The integration of flexible Peltier junctions offers a significant advantage over conventional Joule heating methods by allowing active and reversible heating and cooling cycles. This innovation not only enhances the reliability and longevity of soft robotic applications but also broadens the scope of design possibilities, including the development of agonist-antagonist artificial muscles, grippers with can manipulate through flexion and extension, and an anchor-slip style simple crawler design. Our findings indicate that this approach could lead to more efficient, versatile, and durable robotic systems, marking a significant advancement in the field of soft robotics.

\end{abstract}

\section{Introduction}
The increasing integration of robots into daily life necessitates the development of machines that can safely interact with humans.
Conventional robots, constructed from rigid materials, often lack the necessary compliance for various applications. This limitation has led to significant interest in soft robots and actuators within the robotics community, owing to their inherent flexibility and adaptability \cite{colgate2008safety,de_santis_atlas_2008,schutz2016rrlab, pratt_stiffness_1997}. Soft robots have been explored in diverse domains, including wearable technologies, medical robotics, and more, leveraging various actuation methods such as fluidic actuation, which is noted for its low fabrication costs, lightweight, and high dexterity \cite{marchese_recipe_2015}.

    \begin{figure*}[ht]
    	\vspace{0.25cm}	\centering
        \hspace{-0.0cm}
        \includegraphics[width=0.9\textwidth]{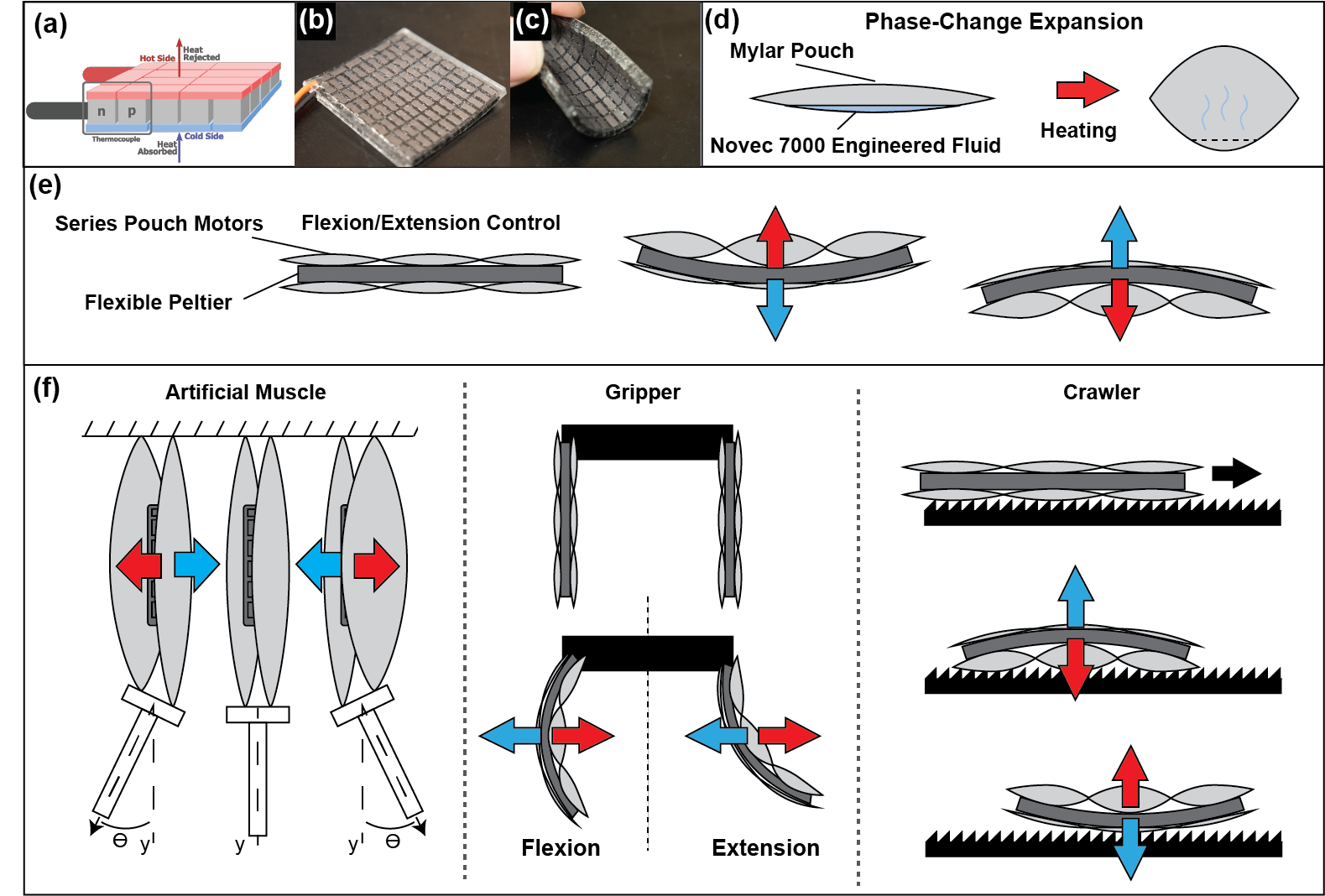}
    
    	\caption {a) Cutaway of a typical Peltier and b) operation principle of Peltier with one p-type and one n-type thermoelement.c) TEGWay Peltier \cite{Tegway} showing size (40x40mm) and flexibility. d) Singular mylar pouch motor filled with Novec 7000 for phase-change expansion. e) Series pouch motors attached to flexible Peltier for reversible flexion and extension. f) Agonist-antagonist configuration of muscle, showing flexion and extension based on heating direction, gripper showing capabilities of reversible flexion and extension, and crawler demonstrating locomotion.}
    	\label{abstract}
    	\vspace{-0.15cm}
    \end{figure*}

The human body achieves complex motions through the simple action of skeletal muscles that contract to produce movement\cite{zajac_muscle_1989}. This is exemplified in the agonist-antagonist relationship seen in bodily movements, such as the biceps and triceps during elbow flexion. Inspired by this, researchers in soft robotics have endeavored to replicate these muscular actions through 'artificial muscles', aiming to achieve diverse motions tailored to specific tasks\cite{tondu_modelling_2012}. However, the challenge remains in finding a functional stimulus that naturally incorporates an agonist-antagonist design\cite{della_santina_soft_2021}.

Addressing this challenge, we propose the innovative integration of flexible thermoelectric devices, specifically Peltier junctions, with soft actuators to mimic the agonist-antagonist mechanism. Unlike traditional heating methods that rely on inefficient Joule heating with passive cooling, Peltier junctions offer a reversible stimulus, allowing for active heating and cooling cycles essential for the proposed agonist-antagonist actuation system.

Phase-change soft actuators represent a compelling domain in the field of soft robotics, renowned for their substantial force output and modest power demands\cite{Miriyev2017-ps}. These actuators typically consist of phase-change fluids integrated in silicone elastomers during the curing process, that can undergo a reversible change in shape in response to thermal variations\cite{decroly_optimization_2021}. Despite the considerable energy density these materials afford, their practical application is often limited by very low bandwidth of their actuation dynamics and the suboptimal efficiency of the heating stimulus (typically Joule heating), lacking direct cooling processes, which necessitates the exploration of more efficient alternatives for rapid actuation\cite{long_latest_2022}.


Peltier junctions represent a frontier in soft robotics, promising to address the inherent limitations of current thermo-active actuators. While thermoelectric devices have found applications in soft robotics for actuation\cite{Yoon2023-jy,Uramune2022-iv}, energy harvesting\cite{zadan_liquid_2022}, and thermal management\cite{greco_liquid_2013}, their integration for actuation purposes remains relatively unexplored. Traditional implementations have predominantly attached actuating units to only one side of the thermoelectric device, relegating the other side to passive roles such as attachment to heatsinks or other thermal management materials. This conventional approach limits the dynamic potential and efficiency of soft actuators.

Furthermore, the prevalent use of stretchable silicone-based materials in soft robotics, such as phase-change composites, has encountered challenges, including leakage and the degradation of phase change fluids necessitated for actuation\cite{exley_utilizing_2023}. These issues highlight the need for innovative solutions that can ensure longevity and reliability in soft robotic applications. Pouch motors\cite{niiyama_pouch_2014,Uramune2022-iv}, which utilize flexible but not stretchable materials, can then be employed to enable geometric modeling without the drawbacks of fluid leakage or the need for fluid 'rejuvenation'.

Our proposed methodology diverges from traditional practices by integrating flexible Peltier junctions in a manner that leverages both sides of the device, embodying the agonist-antagonist principle intrinsic to muscle movements in living organisms. This dual-sided approach not only enhances speed but also amplifies the actuator's range of motion and functionality, allowing for more complex and nuanced control mechanisms.

Through modifying a basic pouch motor design \cite{Uramune2022-iv,li_bio-inspired_2022} to feature multiple channels instead of a single central cavity and incorporating flexible Peltier junctions for agonist-antagonist applications, we can showcase the capability of this technology to propel soft robotics towards enhanced efficiency, reliability, and complexity in actuation systems.

The paper is structured as follows: Section II introduces the design and modeling of the pouch motor. In Section III, experimental testing and analysis of temperature data are presented, along with manufacturing details. Section IV explores potential applications including artificial muscles, grippers, and locomotion. Finally, Section V concludes the paper and outlines future research directions.

\section{Series Pouch Motors}
\subsection{Pouch Design}

In the design of our pouch motors, we utilize thin films that are heat-sealed to create multiple cavities. Unlike conventional designs that typically feature a single, large cavity, our approach segments the pouch into a series of interconnected cavities through narrow channels. By constructing thin channels between the cavities, the pouch is designed to naturally buckle upon volume expansion, resulting in a smooth curvature. This inherent buckling characteristic is key when attached to the flexible Peltier, which serves as a strain-limiting layer.



\subsection{Theoretical Modelling}

\begin{figure}[ht] 
\centering 
\vspace{0.25cm}

\begin{subfigure}[ht]{0.4\textwidth}
    \includegraphics[width=.6\textwidth]{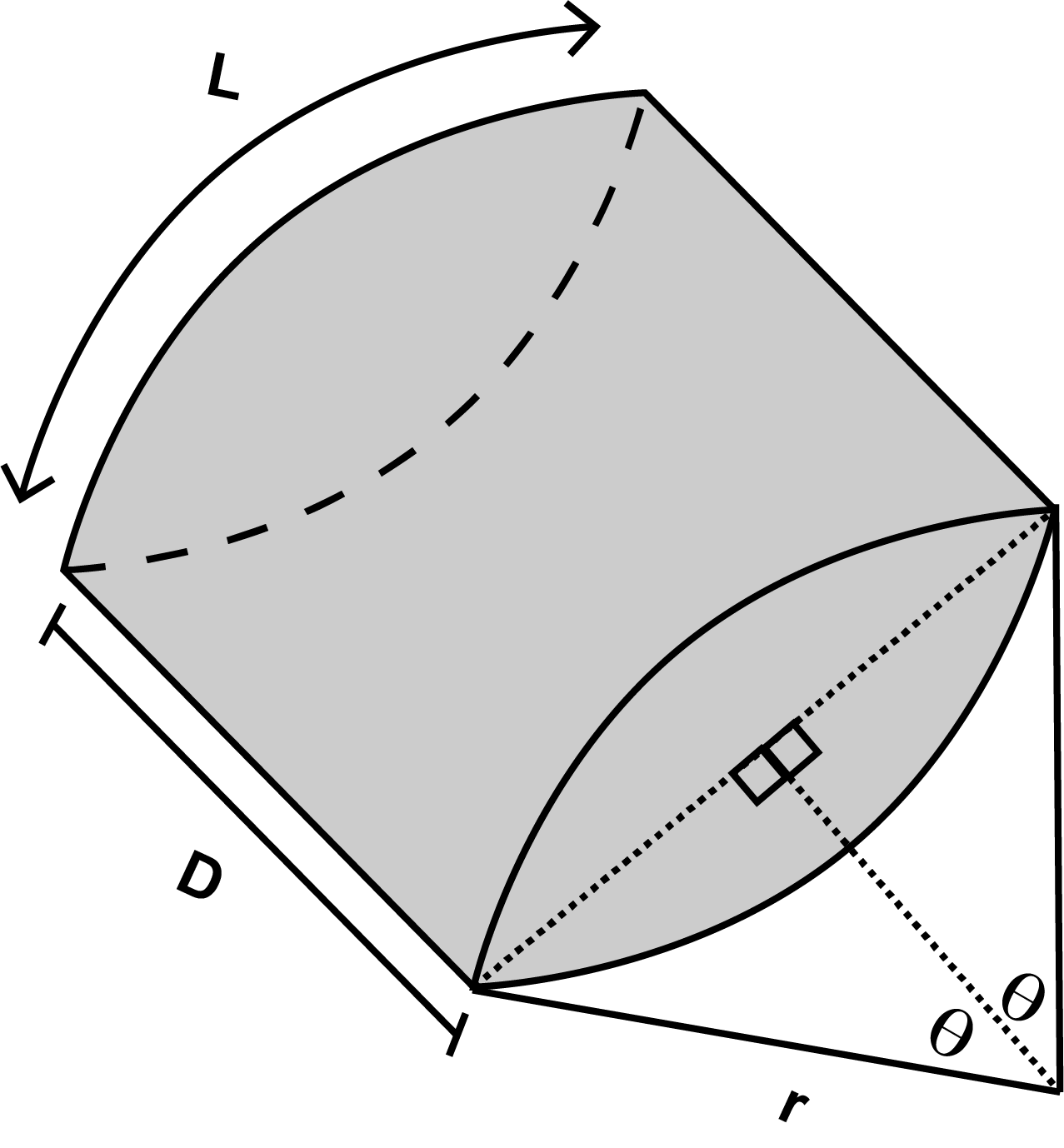}
    \caption{Single cavity}
    \label{fig:modelindividual}
\end{subfigure}
\hfill 

\begin{subfigure}[ht]{0.35\textwidth}
    \includegraphics[width=.6\textwidth]{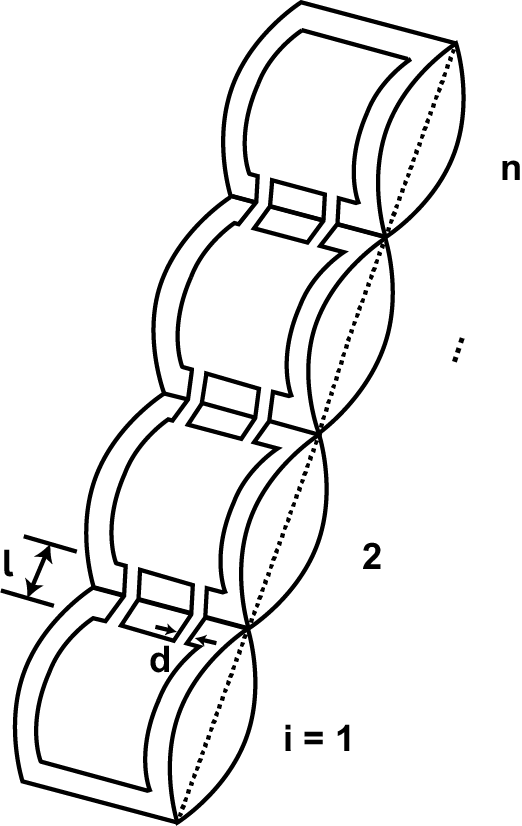}
    \caption{Cavities in series}
    \label{fig:modelseries}
\end{subfigure}
\hfill 

\caption{Geometric model for the Agonist-Antagonist pouch motors}
\label{fig:modeling} 
\end{figure}

We calculated the optimal amount of Novec 7000 inside a liquid-to-gas phase change actuator by using the state equation and theoretical model,  with reference to equations derived by Niiyama et al \cite{niiyama_pouch_2014}, \cite{niiyama_pouch_2015}. When the pressure of Novec 7000 is reached at the vapor pressure at a certain temperature, the pressure in the pouch does not increase anymore. 
Encapsulating a sufficient quantity of Novec 7000 within a sealed enclosure allows for its vaporization until the internal pressure reaches the saturation pressure, denoted as $P_{vap}$ at temperature T [K]. 
The vapor pressure of Novec 7000 can be calculated using the following equation:

\begin{equation}
P_{\mathrm{vap}} = \mathrm{exp}(-3548.6/T + 22.978),
\end{equation}
where  $P_{vap}$ is the vapor pressure at T [K]. 

At saturation pressure both gas and liquid phases of Novec 7000 coexist. Upon reaching the  equilibrium state, the residual Novec 7000 predominantly functions as a heat sink, increasing its specific heat capacity. Therefore, to achieve  a rapid response it is essential to minimize the quantity of remaining liquid state.

$V_{max}$ represents the maximum theoretical volume of the liquid-to-gas phase change actuator. However, during actuation, the pouch edges deform slightly, causing slight discrepancies between the theoretical model and the real-world behavior. Within the scope of this study, we assume that the theoretical and actual volumes are equivalent.

\begin{equation}
V_{\mathrm{max}} = V \left ({\frac {\pi }{2} }\right) = \frac {L^{2} D}{\pi }.
\label{eq: Maximum Volume }
\end{equation}

Finally, the optimal amount of Novec 7000 to fully inflate $n$ cavities of the liquid-to-gas phase change actuator can be expressed as follows:

\begin{equation}
V_{\text{cavity}} = n \cdot \left( \frac{M \cdot P_{\text{vap}} \cdot L^2 \cdot D}{\rho \cdot R \cdot T \cdot \pi} \right).
\label{eq:Volume of Cavity }
\end{equation}
For the channels that connect each cavity, we can then model them as their own cylindrical pouches and factor in the thickness of the seal $l$ and the width of the channel $d$ for:

\begin{equation}
    V_{\text{channel}} = (n - 1) \cdot 2 \cdot \left( \frac{M \cdot P_{\text{vap}} \cdot l^2 \cdot d}{\rho \cdot R \cdot T \cdot \pi} \right).
    \label{eq: Volume of Channel}
\end{equation}
Combining both of these equations can obtain the total volume required for the series pouch motors.

\begin{equation}
\begin{split}
    V_{\text{total}} = & n \cdot \left( \frac{M \cdot P_{\text{vap}} \cdot L^2 \cdot D}{\rho \cdot R \cdot T \cdot \pi} \right) \\
    & + (n - 1) \cdot 2 \cdot \left( \frac{M \cdot P_{\text{vap}} \cdot l^2 \cdot d}{\rho \cdot R \cdot T \cdot \pi} \right).
\end{split}
\label{eq: Total Volume}
\end{equation}

As the $P_{\text{vap}}$ and $T$ are calculated for 34$^\circ$C, the rest of the terms are related to the material and geometry of the pouches, and can be factored to get:

\begin{equation}
\begin{split}
V_{\text{total, factored}} = & \frac{M \cdot P_{\text{vap}}}{\rho \cdot R \cdot T \cdot \pi} \\
& \cdot \left( n \cdot L^2 \cdot D + 2 \cdot (n - 1) \cdot l^2 \cdot d \right).
\end{split}
\label{eq: Total Volume Factored}
\end{equation}

These geometric models can be seen in Fig. \ref{fig:modeling}. In our study, the inclusion of the volume attributed to the channels does not substantially impact calculations. However, as larger or more channels are incorporated, this addition assumes greater significance.



\subsection{Working Principle}

The actuation mechanism of our proposed phase-change pouch motor actuator rests on two fundamental working principles: (i) the reversible phase-change of a housed fluid and (ii) the Peltier thermal effect for rapid heating and cooling. A unit of actuation comprises a Mylar pouch containing a phase-change material that transits from liquid to gas upon heating above its vaporization threshold, and conversely from gas to liquid upon cooling. This process results in reversible volumetric expansion and contraction within the pouch, enabling actuation.

Traditionally, fluorocarbons with lower boiling points, like perfluorocyclohexane at $51^\circ$C, were preferred for their phase-change properties\cite{kato_development_2016}. However, due to environmental concerns, especially their detrimental effects on the ozone layer, alternative materials like Novec 7000 with a vaporization temperature of $34^\circ$C have been identified as viable and eco-friendly substitutes.

A Peltier device (Fig. \ref{abstract}a), also known as a thermoelectric cooler, utilizes the Peltier effect to enact temperature control. When an electrical current is passed through a junction of two dissimilar conductors, heat is either absorbed or released at the junction. This effect is leveraged to precisely control the temperature of the phase-change material within the pouch, as delineated in the following equations:

\begin{equation}
    \dot{Q}_c = n\left[\alpha T_c I - \frac{1}{2}I^2 R - K(T_h - T_c)\right],
\end{equation}

\begin{equation}
    \dot{Q}_h = n\left[\alpha T_h I + \frac{1}{2}I^2 R + K(T_h - T_c)\right].
\end{equation}

Here, $\dot{Q}_c$ and $\dot{Q}_h$ represent the cooling and heating power at the cold and hot junctions, respectively, with $n$ denoting the number of thermocouples, $I$ the current, $\alpha$ the Seebeck coefficient, $R$ the internal resistance, and $K$ the thermal conductivity. Off-the-shelf thermoelectric cooling (TEC) devices are characterized by their maximum temperature difference ($\Delta T$) and power ($Q_{\text{max}}$). In this study, we utilized commercially available flexible Peltiers (TEGWAY, Korea) which notably forgo the conventional rigid ceramic structure in favor of a pliable elastomer embedded with organic filler. This innovative design lends the devices an enhanced flexibility and thermal conductance crucial for the dynamic requirements of our phase-change pouch motor actuators.

The integrated Peltier elements within the actuator's architecture facilitate the rapid transition of the phase-change material between states, thus empowering the actuator with swift response times and high energy efficiency, essential for advanced soft robotic applications.

\section{Fabrication and Analysis}

\subsection{Temperature Data Analysis}

We assess the performance characteristics of flexible Peltier devices. These devices vary in size, but for the purpose of temperature data analysis, a standard dimension of 40 mm by 40 mm was utilized. The primary advantage of flexible Peltiers lies in their ability to generate heat concurrently with cooling capabilities when currents are reversed. Unlike conventional heaters, which solely produce heat and rely on ambient conditions for cooling, Peltier devices actively facilitate both heating and cooling processes. This dual functionality renders them well-suited for use in physically interactive human-robot interaction (pHRI) scenarios within dynamic environments. The evaluation of Peltier elements was conducted across three distinct power settings: 2.5 W, 5 W, and 7 W, with five trials performed at each power setting.

In connection with the power supply through alligator clamps rests a fabricated current inversion switch where it harnesses the current inversion feature to control heating and cooling characteristics. Additionally, the switch also allows the swap from heating to cooling at selected times. From there, the switch is directly connected to a flexible 40 mm x 40 mm Peltier. All of the setup rests under a camera mount housing the FLIR C5 Compact Thermal Imaging Camera (Teledyne FLIR: OR, USA) powered through internal battery overlooking testing environment. To ensure consistent and optimal contact of the heating element at all times, a single sheet of Mylar, heat-sealable side up, was attached onto the flexible Peltier and Joule heater via a layer of graphite applied onto polyethylene. 

\begin{figure}[ht]
    \centering
    \vspace{0.25cm}
    \includegraphics[width=.95\columnwidth]{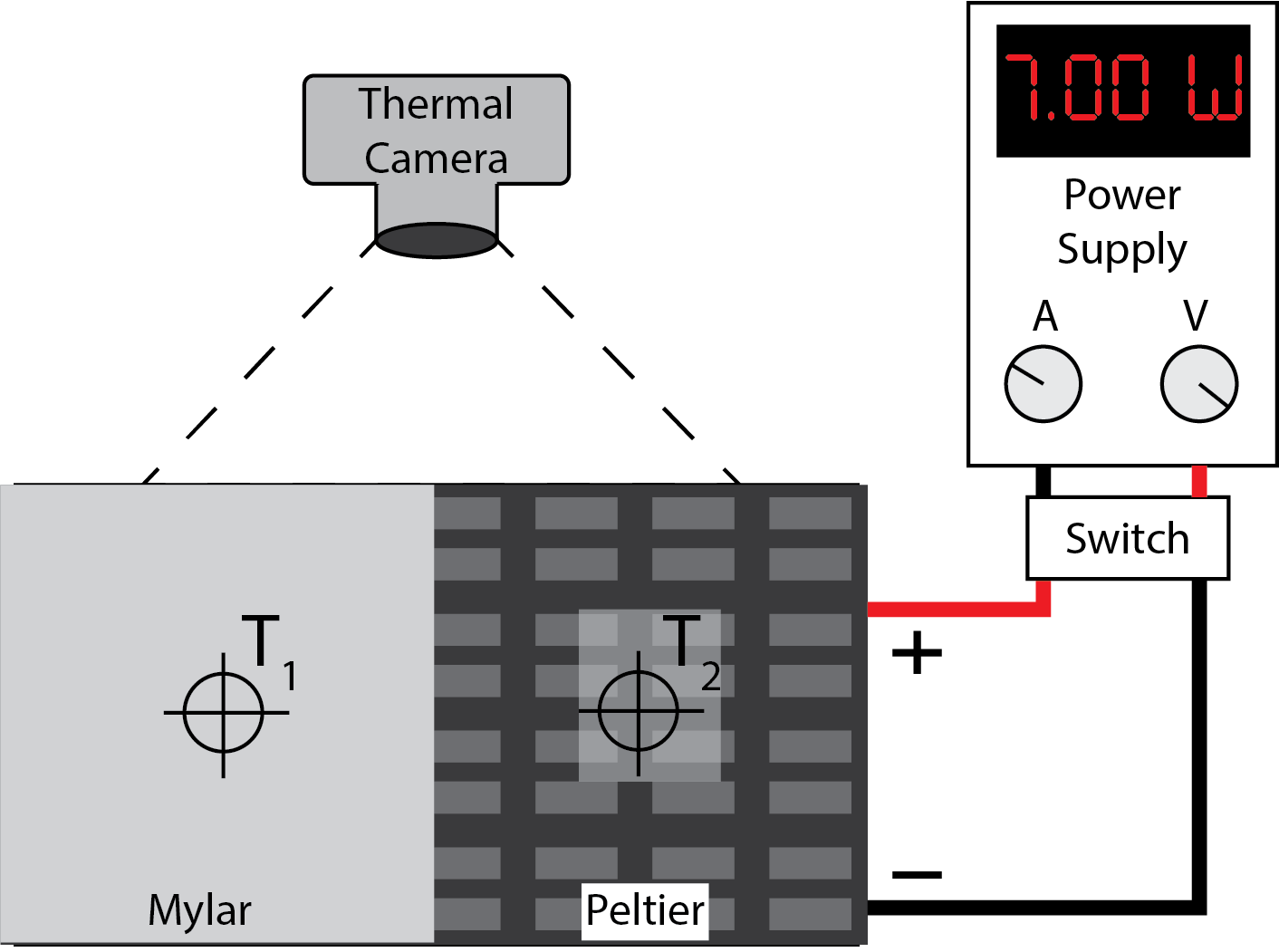}
    \caption{Experimental setup for collecting Peltier and Mylar temperature values. In the case of flexible polyimide heater, the Peltier is substituted.}
    \label{fig:tempsetup}
\end{figure}



Through testing of the flexible Peltier element, the heating source models distinct behavior at the 3 varying power settings. The data testing aimed to stop at the same temperature for all the Peltiers, at 70$^\circ$C. It can be observed that at 2.5 W, the Peltier takes significantly longer time to reach the desired temperature of 70$^\circ$C due to its low current input. At 5 W and 7 W, distinct behaviors emerge as the surface temperature of both the Mylar and Peltier exhibit exponential increases, increasing at a faster rate than the 2.5 W powered source. However, their trajectories diverge after 10 seconds. Following this initial period, the Peltier operating at 5 watts experiences a dramatic deceleration, whereas the 7 W counterpart maintains its ascent towards 70$^\circ$C. Hypothesizing that the Peltier powered with higher wattage (5 W and 7 W) will experience rapid decent towards ambient temperature, this behavior can be observed. The 5 W and 7 W heater estimates to take up to 10 seconds to reach the ambient temperature of 35$^\circ$C while the 2.5 W flexible Peltier shows much slower cooling time up to 30 seconds due to the fact that the slow heating and slow heating may be caused by the shortage of power provided to the heater.


\subsection{Fabrication}


    \begin{figure*}[ht]
    	\vspace{0.25cm}	\centering
        \hspace{-0.0cm}
        \includegraphics[width=0.95\textwidth]{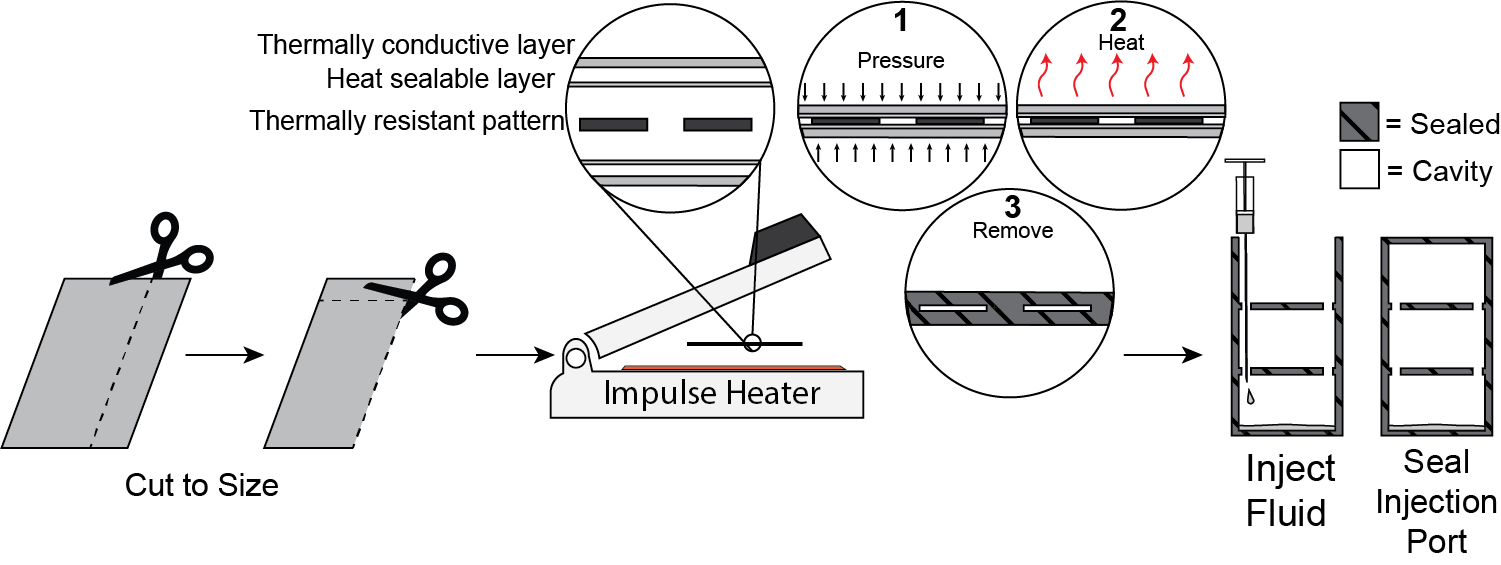}
    
    	\caption {For the creation of the pouch motors, mylar films are cut to size, and then boundaries are sealed using an impulse heater. Thin (width: 2 mm; thickness: 150 $\mu$m) Nichrome ribbon wire was placed into the pouch to create areas that are thermally resistant according to designed internal cavity geometry. Once placed, sealing will only occur where wire is absent. After sealing, wire is removed and corresponding Novec 7000 volume is injected using a 22 gauge needle, and injection port is sealed.}
    	\label{methodology}
    	\vspace{-0.15cm}
    \end{figure*}
    
Fig. \ref{methodology} shows the entire manufacturing process of liquid to gas phase change actuator. For the creation of the pouch motors, mylar films are cut to size, and then boundaries are sealed using an impulse heater. Thin (width: 2 mm; thickness: 150 $\mu$m) Nichrome ribbon wire was placed into the pouch to create areas that are thermally resistant according to designed internal cavity geometry. Once placed, sealing will only occur where wire is absent.  After sealing, wire is removed and corresponding Novec 7000 volume is injected using a 22 gauge needle, and injection port is sealed .this makes sure that the vaporized fluid can only flow between individual pouches at these points. Allowing better control of fluid flow between the pouches.

To manufacture the pouches, the Mylar sheets are cut to specified dimensions of each application. The pouch is then sealed on all sides except for one designated opening. Next, nichrome strips are inserted on either side of the sealed pouch so that it aligns on the sealed seams.  The pouch is then finally sealed horizontally, creating a series arrangement of the individual pouches. Notably, the nichrome wires prevent the Mylar from sealing at the contact points of the nichrome wire, 
After sealing the nichrome wire is removed and  Novec 7000 is injected using a syringe. 

The syringe was inserted to the edge of the last pouch and fluid injected to ensure no fluid gets vaporized during sealing.  Finally, the remaining opening is sealed to complete the series pouch motor unit.

\begin{figure}
    \centering
    \includegraphics{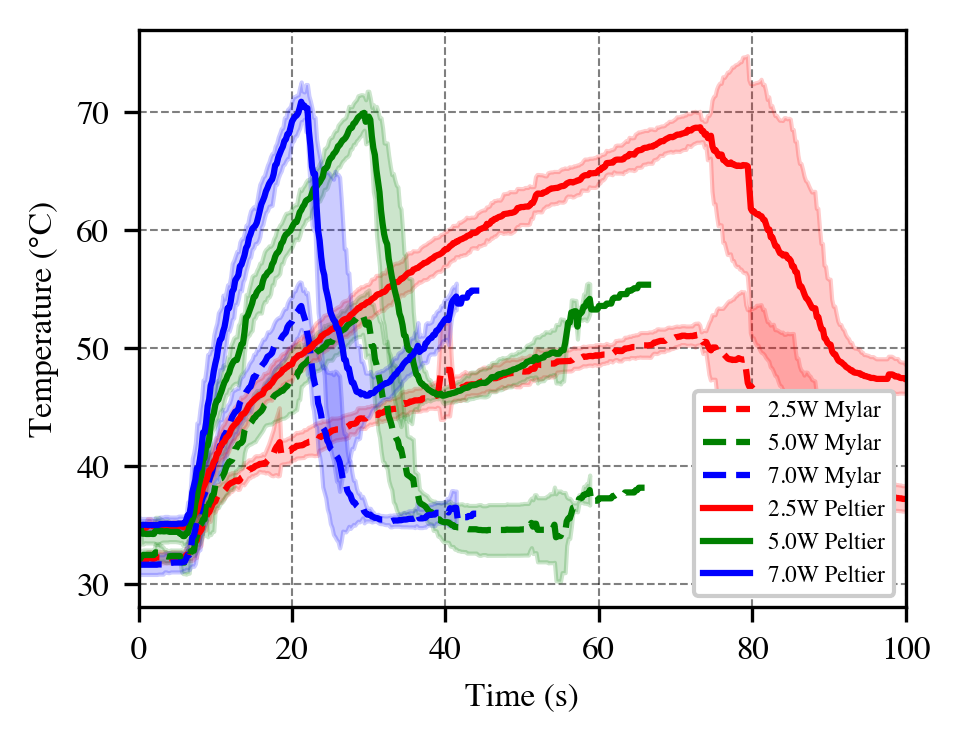}
    \caption{Temperature of inner side of mylar and surface of Peltier collected with the setup from Fig. \ref{fig:tempsetup}}
    \label{fig:temperaturetest}
\end{figure}

\section{Agonist-Antagonist Applications}
In leveraging the unique capabilities of the flexible Peltier junction for thermo-active actuation, we have conceptualized and developed three distinct applications that embody bioinspired agonist-antagonist designs. These applications are intricately designed to mimic natural biological systems, utilizing the dual heating and cooling functions of the Peltier devices to achieve dynamic and responsive actuation.

\subsection{Artificial Muscle}

\begin{figure}
    \centering
    \vspace{0.175cm}
    \includegraphics[width=0.45\textwidth]{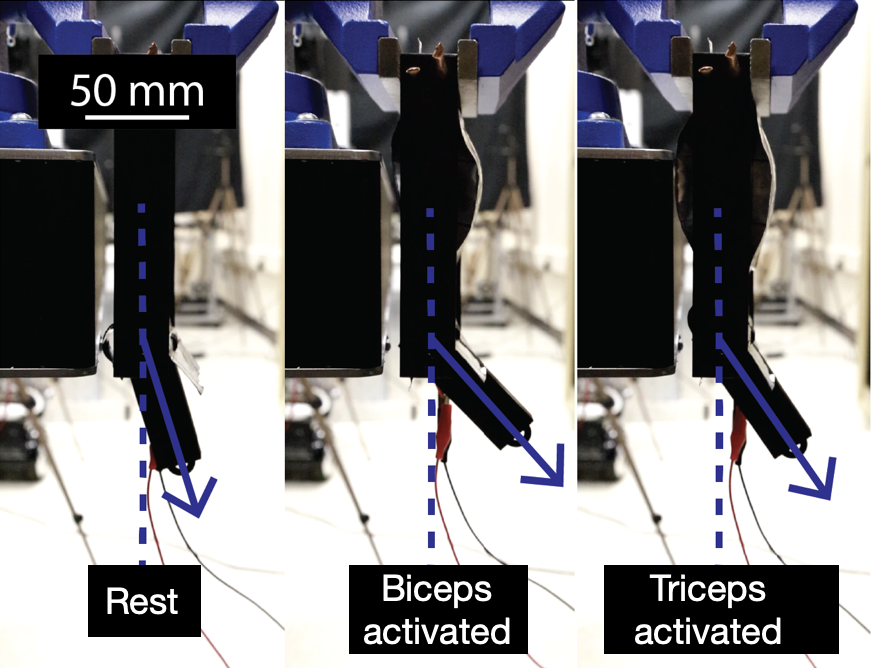}
    \caption{Muscle application depicting biceps and triceps activation, contributing to both flexion/extension and stiffness regulation, with heating and cooling from Peltier.}
    \label{fig:muscle}
\end{figure}

The first application is a two-muscle joint system, which closely mirrors the functionality of human joints. 
By employing a Peltier junction between two pouches as artificial muscles (biceps and triceps), both flexion and extension of the output link can be achieved through the Peltier effect, which facilitates the transfer of heat from one side to the other (Fig. \ref{fig:muscle}). This mechanism allows for the activation of one pouch muscle while the other remains uninflated, enabling the extension or flexion of the link depending on which muscle is activated. Activation of both muscles, achieved by energizing the Peltier (heating both sides), results in stiffening of the link. Thus, a potential operational approach involves rotating the output link to a desired angle by activating only one muscle (depending on the direction of current flow), and subsequently deactivating the Peltier to enhance the stiffness of the output link at the desired angle, until ambient temperature regulation takes effect. This approach necessitates precise control over the timing, direction, and magnitude of the electric current supplied to the Peltier. Each muscle has a weight of 6.38 g, with an additional 9.75 g for the Peltier



\subsection{Gripper}

\begin{figure}
    \centering
    \includegraphics[width=.425\textwidth]{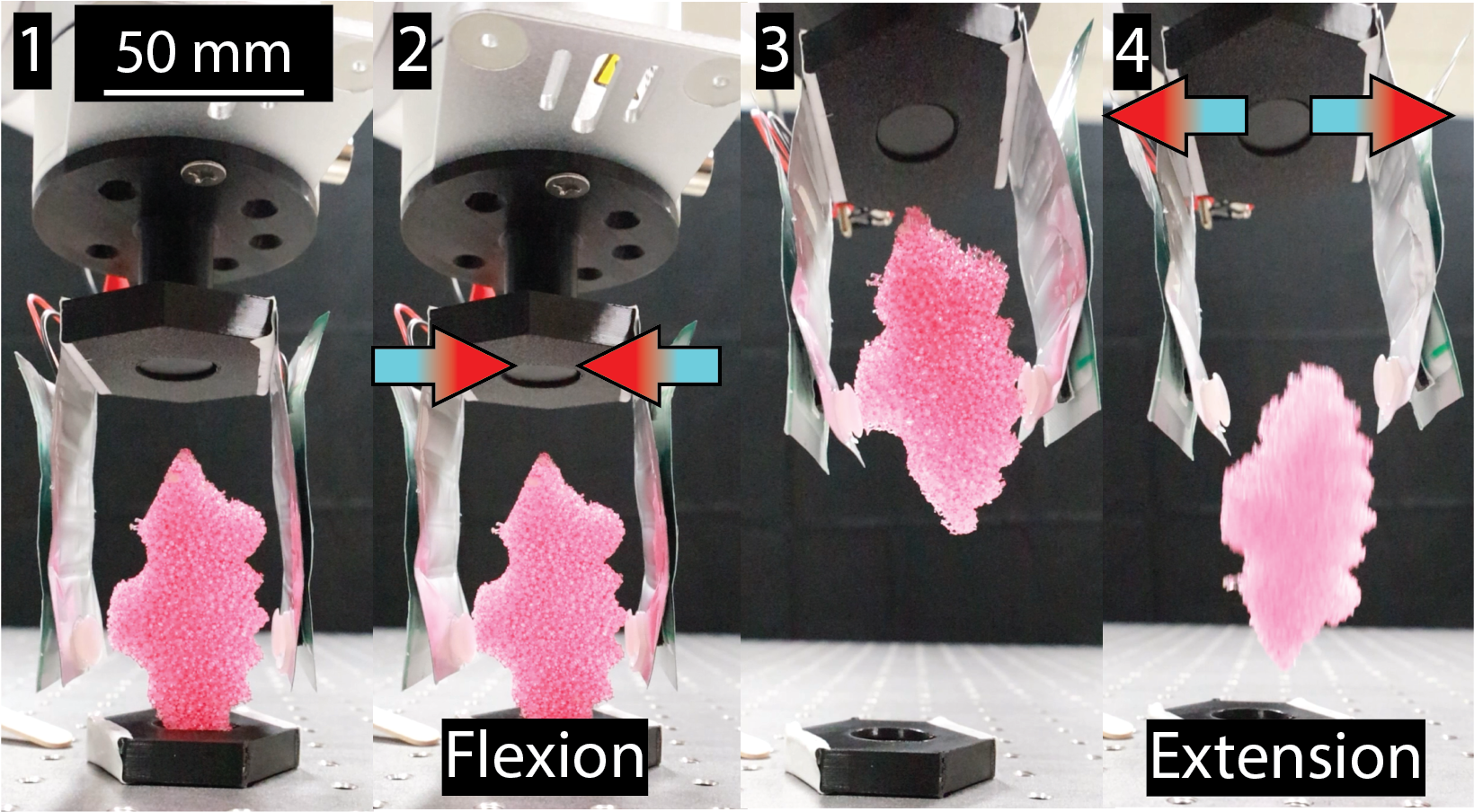}
    \caption{Agonist-antagonist gripper design picking up irregular-shaped sponge (m = 1.71 g) using flexion.}
    \label{fig:gripper}
\end{figure}

\begin{figure}
    \centering
    \includegraphics[width=.45\textwidth]{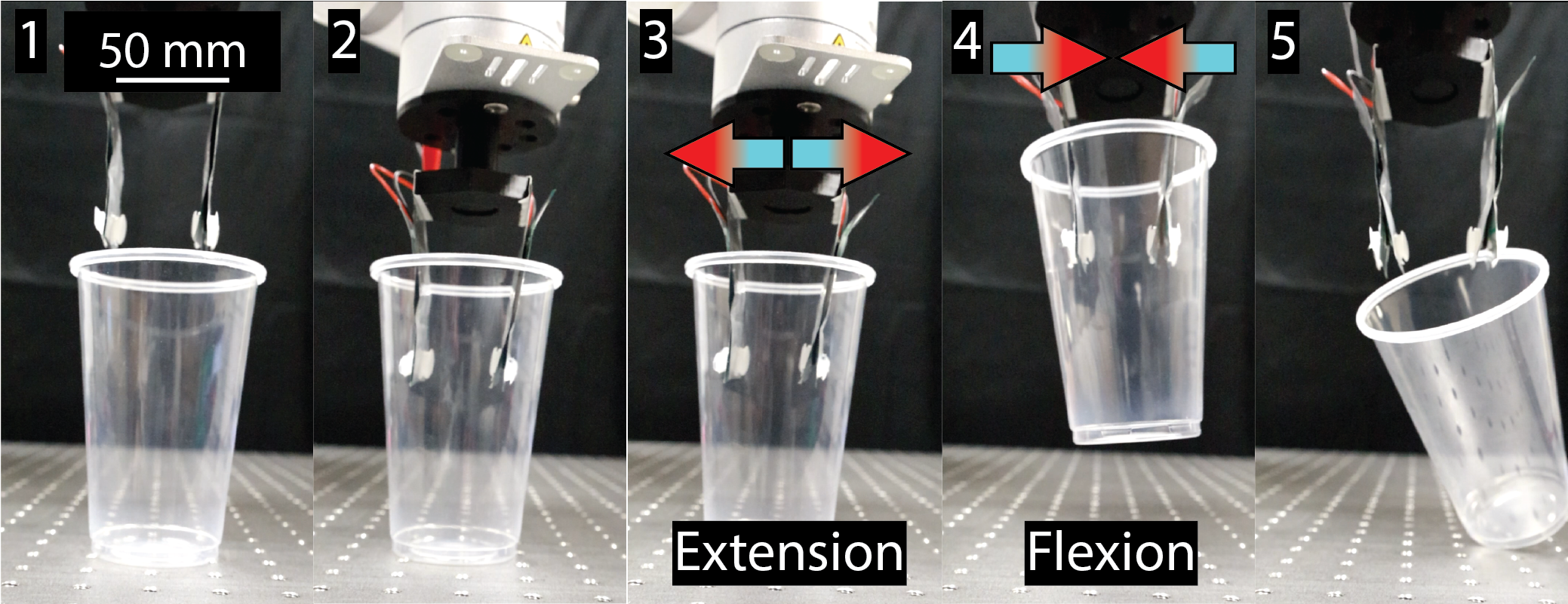}
    \caption{Agonist-antagonist gripper design picking up cup (m = 9.0 g) using extension.}
    \label{fig:gripper2}
\end{figure}

Following the two-muscle joint system, we explored the development of a soft robotic gripper. This gripper is designed to adjust its grasp both inwards (Fig. \ref{fig:gripper}) and outwards (Fig. \ref{fig:gripper2}). The designed end effector was attached to an xArm 5 (UFactory; Shenzhen, China). The weight of each gripper finger is 4.85 g, with an additional 2.55 g for the Peltier. The gripper pouches have incorporated silicone (Ecoflex 00-50, SmoothOn) pads at point of contact. 

It's important to note that unlike other grippers employing soft actuation, our design does not incorporate any rigid elements. Typically, rigid structures are utilized to constrain the lateral movement of actuators, enhancing their gripping force on objects and enabling them to hold heavier items. This is because soft actuators, composed of inherently soft materials, can only exert pulling force, and their structure would bend in response to a pushing force. Nevertheless, due to agonist-antagonist design of our gripper, we were able to grasp lighter objects, including those with irregular shapes, by incorporating friction-based materials at the fingertips.

\subsection{Locomotion}

\begin{figure}
    \centering
    \vspace{0.175cm}
    \includegraphics[width=0.45\textwidth]{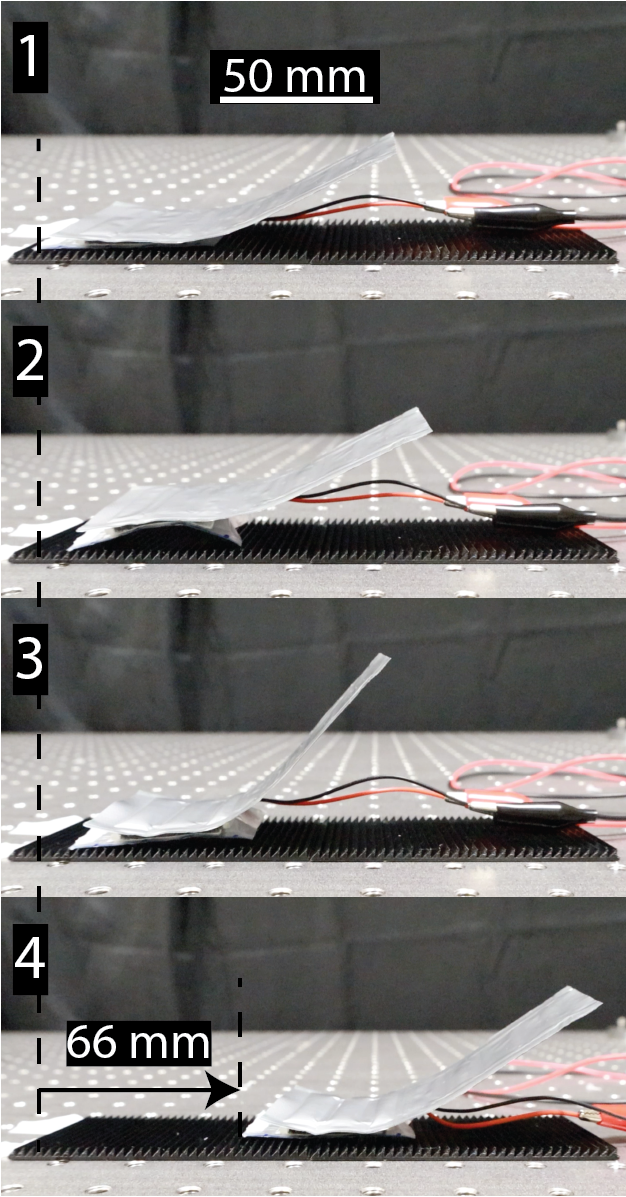}
    \caption{Crawler locomotion design placed on an asymmetric sawtooth 3d-printed substrate. Location 1: 0 s; Location 2: 18 s; Location 3: 26 s; Location 4: 871 s.}
    \label{fig:locomotion}
\end{figure}

Lastly, the locomotion application represents an implementation of bioinspired design principles, achieving locomotion through coordinated actuation. This design draws inspiration from an inchworm, keeping two points of friction on the surface as it bends (Fig. \ref{fig:locomotion}). Using this anchor-slip style of movement enables a repeatable gait as the current through the Peltier is alternated \cite{duggan_inchworm-inspired_2019}. This is tested on an asymmetric sawtooth 3D printed substrate. 
It's worth noting that by optimizing the duration of Peltier operation, we can ensure consistent and high-quality movement. This approach allows us to effectively manage temperature fluctuations and maintain optimal performance throughout operation. 
Incorporating some hysteresis control (i.e., alternating between high and low currents) could increase the overall speed. The crawler showcases the potential of thermo-active soft actuators in creating more autonomous and versatile robotic systems capable of navigating diverse environments.

\section{Conclusion}
The conventional reliance on stretchable elastomers for encapsulating phase-change fluids has posed significant challenges, particularly in terms of maintaining the fluid's integrity within the actuator. This research proposes a shift towards the use of purely flexible, non-stretchable materials, such as Mylar, in the construction of phase-change actuators. This approach implements flexible Peltiers and emphasizes agonist-antagonist designs addresses the limitations of elastomers by minimizing the thermal path required for actuation, thereby ensuring a more efficient and controlled thermal response. The use of non-stretchable materials ensures that the phase-change fluid remains in close proximity to the planar surfaces during rest, enhancing the actuator's performance and reliability.


\addtolength{\textheight}{0cm}






\bibliographystyle{IEEEtran}
\bibliography{biblio.bib}

\end{document}